\documentclass{tlp}
\usepackage{amssymb}
\usepackage{amsmath}
\usepackage{mathrsfs}
\usepackage{color}
\usepackage{url}

\newcommand{\la}{\leftarrow}

\newcommand{\PP}{\mathcal{P}}
\newcommand{\es}{\emptyset}

\newcommand{\nwf}{\textcolor{black}}
\newcommand{\nw}{\textcolor{black}}

\newtheorem{definition}{Definition} 
\newtheorem{example}{Example} 
\newtheorem{remark}{Remark} 

\newtheorem{proposition}{Proposition}
\newtheorem{lemma}{Lemma}

\submitted{9 December 2014}
\revised{2 July 2015}
\accepted{14 December 2015}

\begin{document}

\long\def\comment#1{}

\title[Solving SMP using ASP]{Solving Stable Matching Problems\\ using Answer Set Programming}

\author[S. {De Clercq} et al.]
{SOFIE DE CLERCQ\\
Dept.\ of Applied Mathematics, Computer Science \& Statistics, 
Ghent University, Belgium\\
\email{SofieR.DeClercq@ugent.be}
\and STEVEN SCHOCKAERT\\
School of Computer Science \& Informatics, 
Cardiff University, UK\\
\email{SchockaertS1@cardiff.ac.uk}
\and MARTINE DE COCK\\
Center for Data Science, 
UW Tacoma, US\\
Dept.\ of Applied Mathematics, Computer Science \& Statistics, 
Ghent University, Belgium\\
\email{MDeCock@u.washington.edu}
\and ANN NOWE\\
Computational Modeling Lab, 
Vrije Universiteit Brussel, Belgium\\
\email{ANowe@vub.ac.be}
}

\pagerange{\pageref{firstpage}--\pageref{lastpage}}
\volume{\textbf{10} (3):}
\jdate{March 2002}
\setcounter{page}{1}
\pubyear{2002}

\maketitle

\label{firstpage}

\begin{abstract}
\nwf{Since the introduction of the stable marriage problem (SMP) by Gale and Shapley (1962), several variants and extensions have been investigated. While this variety is useful to widen the application potential, each variant requires a new algorithm for finding the stable matchings.
To address this issue, we propose an encoding of the SMP using answer set programming (ASP), which can straightforwardly be adapted and extended to suit the needs of specific applications.
The use of ASP also means that we can take advantage of highly efficient off-the-shelf solvers.
To illustrate the flexibility of our approach, we show how our ASP encoding naturally allows us to select optimal stable matchings, i.e.\ matchings that are optimal according to some user-specified criterion.
To the best of our knowledge, our encoding offers the first exact implementation to find sex-equal, minimum regret, egalitarian or maximum cardinality stable matchings for SMP instances in which individuals may designate unacceptable partners and ties between preferences are allowed.
}

\end{abstract}

\section{Introduction}
\nwf{The stable marriage problem (SMP) is a well-known matching problem introduced by Gale and Shapley (1962). The input of an SMP instance consists of (i) a set of $n$ men and $n$ women, (ii) for each man a ranking of the women as preferred partners, and (iii) for each woman a ranking of the men as preferred partners. 
A \textit{blocking pair} of an SMP instance consists of a man and a woman who are in different marriages but both prefer each other to their actual partners. Given the problem, one can compute a \textit{stable matching} or \textit{stable set of marriages}, which is defined as a set of $n$ couples (marriages) such that there are no blocking pairs. Due to its practical relevance, countless variants on the SMP have been investigated, enabling a wider range of applications. \nw{Examples of such applications include} the kidney-exchange problem~\cite{SMPIrv07}\nw{, which matches donors in incompatible donor-recipient pairs to compatible recipients in other incompatible pairs and vice versa,} and the hospital-resident problem~\cite{SMPMan02}\nw{, which matches residents to the free positions in hospitals}. In 2012, Roth and Shapley won the Nobel Prize for Economics for their theory of stable allocations and the practice of market design, which directly resulted from an application of the SMP.}

\nwf{In the literature, typically each time a new variant or generalization of the SMP or a different optimality criterion is considered, a new algorithm is developed; see e.g. \cite{SMPGus87,SMPIrv87,SMPMc12}, or~\cite{SMPMan13} for an overview. In this paper, we propose to use answer set programming (ASP) as a general vehicle for modeling a large class of extensions and variations of the SMP. We show how ASP encodings can be used to compute stable matchings, and how this encoding can be extended to compute optimal stable matchings.
Although the SMP has been widely investigated, and efficient approximation or exact algorithms are available for several of its variants~\cite{SMPIw10,SMPMc12}, to the best of our knowledge, our encoding offers the first exact implementation to find sex-equal, minimum regret, egalitarian or maximum cardinality stable matchings for SMP instances with unacceptable partners and ties (see below).}

\nwf{In this paper, we will consider two well-known adaptions of the SMP. First, we will consider problem instances in which every person can specify a set of unacceptable partners.
The second alteration consists of allowing ties in the preferences, i.e.\ one can be indifferent between some possible partners. In the literature\nw{,} the SMP variant with unacceptable partners \nw{-- or, equivalently, with incomplete preference lists --} is abbreviated as SMI. \nw{The variant with ties is denoted as SMT and the variant which allows both extensions as SMTI.} 
Note that the original SMP is a special case of the SMTI, i.e.\ the set of unacceptable partners is empty for each man and woman, and there are no ties. Therefore, our paper focusses on the SMTI variant, as it is the most general one.}

\nwf{Another way to generalize the SMP is by introducing optimality criteria for stable matchings. This is motivated by the fact that, if multiple stable matchings exist, some may be more interesting than others. In this paper, we focus on sex-equality, minimum regret, egalitarity and maximum cardinality of the stable matchings, as these are commonly investigated optimality criteria in the matching literature. 
Note, however, that there exist several other optimality criteria in the context of matchings, such as \textit{popularity}~\cite{SMPGard75}, \textit{Pareto optimality}~\cite{SMPGale85,SMPRoth90} or profile-based notions such as rank-maximum~\cite{SMPIrv03}, greedy maximum and generous maximum~\cite{SMPIrv03}.
As this list is non-exhaustive, we refer the interested reader to~\cite{SMPMan13} for an overview.
}

\nwf{The structure of the paper is as follows. First we give some background about the SMP and ASP in Section~\ref{sec:background}. Then we introduce our encoding of the SMTI with ASP and prove its correctness in Section~\ref{sec:smpinasp}. To illustrate the flexibility of the approach, we show how it can be used to tackle three-dimensional stable matching problems in Section~\ref{sec:3dsmpinasp}. In~Section \ref{sec:oss}, we first discuss several optimality criteria and then extend the encoding from Section~\ref{sec:smpinasp}, enabling us to find optimal stable matchings. We~show how optimal stable matchings of an SMTI instance can be found by solving the corresponding induced disjunctive ASP program and prove the soundness of our approach.}
\nwf{This paper is an extended version of~\cite{SMPClercq13} and additionally provides detailed examples, complete correctness proofs and an ASP encoding of the three-dimensional stable matching problem.}
\nw{The three-dimensional stable matching problem is very important for practical applications, such as the kidney exchange program~\cite{SMPBi10}.}

\section{Background}\label{sec:background}
\subsection{The Stable Marriage Problem}
To solve the original SMP, Gale and Shapley~\cite{SMPGale62} constructed an iterative algorithm ---known as the Gale-Shapley algorithm, G-S algorithm or deferred-acceptance algorithm--- to compute a particular stable matching of an SMP instance. The algorithm works as follows: in round 1 every man proposes to his first choice of all women. A woman, when being proposed, then rejects all men but her first choice among the subset of men who proposed to her. That first choice becomes her temporary husband. In the next rounds, all rejected men propose to their first choice among the subset of women by whom they were not rejected yet, regardless of whether this woman already has a temporary husband. Each woman, when being proposed, then rejects all men but her first choice among the subset of men who just proposed to her and her temporary mate. This process continues until all women have a husband. This point, when everyone has a partner, is always reached after a polynomial number of steps and the corresponding set of marriages is stable~\cite{SMPGale62}. It should be noted, however, that only one of the potentially exponentially many stable matchings is found in this way. 

The classical SMP can be generalized by (i) allowing men and women to point out unacceptable partners, i.e.\ exclude them from their preference list and (ii) dropping the restriction that the number of men $n$ equals the number of women $p$. In this variant, men and women can remain single in a stable matching. 
\nwf{Intuitively, one prefers remaining single over being matched with an unacceptable partner.
This variant is also referred to as the SMP with incomplete preference lists, abbreviated as SMI.} A stable matching for an SMI instance always exists and can be found in polynomial time~\cite{SMPRoth90} by a slightly modified G-S algorithm.
\nwf{As we focus on an extension of the SMI, we refer to the online appendix for the formal definitions of the classical SMP and the SMI.}

\nwf{The SMI variant can further be generalized by additionally allowing ties in the preference lists.} For this variant (SMTI) there are several ways to define stability, but we will use the notion of weak stability~\cite{SMPIrv94}.
We denote a set of men as $M=\{m_1,\hdots,m_n\}$ and a set of women $W=\{w_1,\hdots,w_p\}$. A set of marriages or a matching is a collection of man-woman pairs and singles (persons paired to themselves) such that every man and every woman occurs in just one pair.
\begin{definition}[SMTI] \label{def:smti}
An instance of the SMTI is a pair $(S_M,S_W)$ with $S_M = \{\sigma_M^1,\hdots,\sigma_M^n\}$ and $S_W = \{\sigma_W^1,\hdots,\sigma_W^p\}$. For every $i\in\{1,\hdots,n\}$, $\sigma_M^i$ is a list of disjoint subsets of $\{1,\hdots,p\}$. Symmetrically $\sigma_W^i$ is a list of disjoint subsets of $\{1,\hdots,n\}$ for every $i\in\{1,\hdots,p\}$. We call $\sigma_M^i$ and $\sigma_W^i$ the preferences of man $m_i$ and woman $w_i$ respectively and we denote the length of the list $\sigma_M^i$ as $|\sigma_M^i|$. If $k \in \sigma_M^i(j)$, woman $w_k$ is in man $m_i$'s $j^{th}$ most preferred group of women. All the women in that group are equally preferred by $m_i$. The case $k \in \sigma_W^i(j)$ is similar. If there is no $l$ such that $j \in \sigma_M^i(l)$, woman $w_j$ is an unacceptable partner for man $m_i$, and similarly when there is no $l$ such that $j \in \sigma_W^i(l)$. For every $k$ in the set $\sigma_M^i(|\sigma_M^i|)$, man $m_i$ equally prefers staying single to being paired to woman $w_k$, and symmetrically for the preferences of a woman $w_i$. This is the only set in $\sigma_M^i$ that might be empty, and similar for $\sigma_W^i$. 
Man $m$ and woman $w$ form a blocking pair in a set of marriages $S$ if $m$ strictly prefers $w$ to his partner in $S$ and $w$ strictly prefers $m$ to her partner in~$S$. A blocking individual in $S$ is a person who stricly prefers being single to being paired to his partner in $S$.
A~weakly stable matching is a set of marriages without blocking pairs or individuals. 
\end{definition}
A weakly stable matching always exists for an instance of the SMTI and it can be found in polynomial time by arbitrarily breaking the ties~\cite{SMPIwa08}. However, as opposed to the SMI variant, the number of matched persons is no longer constant for every stable matching in this variant. 

We introduce the following notations:
\[
acceptable_M^i = \underbrace{\sigma^i_M(1) \cup \sigma^i_M(2) \cup \ldots \cup \sigma^i_M(|\sigma^i_M|-1)}_{\displaystyle = \mbox{\textit{preferred}}_M^i} \cup \underbrace{\sigma^i_M(|\sigma^i_M|)}_{\displaystyle = neutral_M^i}
\]
Furthermore $unacceptable_M^i = \{1,\hdots,p\} \setminus acceptable_M^i$. We define the ordering $\leq_M^{m_i}$ on $\{w_j\,|\, j \in acceptable_M^i\} \cup \{m_i\}$ as $x \leq_M^{m_i} y$ iff $m_i$ prefers person $x$ at least as much as person $y$. 
\nw{Note that $m_i$ is included in its own preference ordering to encode the possibility of staying single.} 
The strict ordering $<_M^{m_i}$ is defined in the obvious way and analogous notations are used for $\sigma_W^j$.
\begin{example} \label{ex:indif}
Suppose $M=\{m_1$, $m_2\}$, $W=\{w_1$, $w_2$, $w_3$, $w_4\}$ and $S_M = \{ \sigma_M^1$ = $(\{1,3\},\{4\})$, $\sigma_M^2$ = $(\{2,3\},\{\})\}$. Hence man $m_1$ prefers women $w_1$ and $w_3$ to woman $w_4$. There is a tie between woman $w_1$ and $w_3$ as well as between woman $w_4$ and staying single. Woman $w_2$ is unacceptable for man $m_1$. Man $m_2$ prefers woman $w_2$ and $w_3$ to staying single, but finds $w_1$ and $w_4$ unacceptable. It holds that $w_1 <_M^{m_1} m_1$, i.e.~$m_1$ prefers marrying $w_1$ over staying single, \textit{acceptable}$_M^1=\{1,3,4\}$, \textit{preferred}$_M^1=\{1,3\}$, $neutral_M^1= \{4\}$ and $unacceptable_M^1=\{2\}$.
\end{example}

\subsection{Answer Set Programming}
Answer set programming or ASP is a form of declarative programming~\cite{ASPBrew11}. Its transparence, elegance and ability to deal with $\Sigma_2^P$-complete problems make it an attractive method for solving combinatorial search and optimization problems. An ASP program is a finite collection of first-order rules
\begin{align*}
A_1 \vee \hdots \vee A_k \la B_1,\hdots,B_m,not \,C_1,\hdots, not \, C_n
\end{align*}
\nw{where $A_1, \hdots, A_k, B_1,\hdots,B_m,C_1,\hdots,C_n$ are predicates, possibly negated by $\neg$, and $not$ is the negation-as-failure operator, whose meaning is explained below.}
The semantics are defined by the \textit{ground version} of the program, consisting of all ground instantiations of the rules w.r.t.\ the constants that appear in it (see e.g. \cite{ASPBrew11} for a good overview). This grounded program is a propositional ASP program. The building blocks of these programs are \textit{atoms}, \textit{literals} and \textit{rules}. The most elementary are \textit{atoms}, which are propositional variables that can be true or false. A \textit{literal} is an atom or a negated atom\nw{, denoted with $\neg$}. Beside strong negation, ASP uses a special kind of negation, namely \textit{negation-as-failure} (naf), denoted with `$not$'. For a literal $a$ we call `$not \, a$' the naf-literal associated with $a$. The \textit{extended literals} consist of all literals and their associated naf-literals. A~\textit{disjunctive rule} has the following form
\begin{align*}
a_1 \vee \hdots \vee a_k \la b_1,\hdots,b_m,not \,c_1,\hdots, not \, c_n
\end{align*}
where $a_1, \hdots, a_k, b_1,\hdots,b_m,c_1,\hdots,c_n$ are literals from a fixed set $\mathcal{L}$, determined by a fixed set $\mathcal{A}$ of atoms. We call $a_1 \vee \hdots \vee a_k$ the head of the rule while the set of extended literals $b_1,\hdots,b_m,not \,c_1,\hdots$, $not \, c_n$ is called the \textit{body}.
The rule above intuitively encodes that $a_1$, $a_2$, $\hdots$ or $a_k$ is true when we have evidence that $b_1,\hdots,b_m$ are true and we have no evidence that at least one of $c_1,\hdots,c_n$ is true. When a rule has an empty body, we call it a \textit{fact}; when the head is empty, we speak of a \textit{constraint}. A rule without occurrences of $not$ is called a \textit{simple disjunctive rule}. A \textit{simple disjunctive} ASP program is a finite collection of simple disjunctive rules and similarly a \textit{disjunctive} ASP program $\PP$ is a finite collection of disjunctive rules. If each rule head consists of at most one literal, we speak of a \textit{normal} ASP program. 

We define an \textit{interpretation} $I$ of a disjunctive ASP program $\PP$ as a subset of $\mathcal{L}$. An interpretation $I$ \textit{satisfies} a simple disjunctive rule $a_1 \vee \hdots \vee a_k$ $\la b_1,\hdots,b_m$ when $a_1 \in I$ \nw{or $a_2 \in I$ or} $\hdots$ or $a_k \in I$ or $\{b_1,\hdots,b_m\} \not \subseteq I$. An interpretation which satisfies all rules of a simple disjunctive program is called a \textit{model} of that program. 
An interpretation $I$ is an \textit{answer set} of a simple disjunctive program $\PP$ iff it is a minimal model of $\PP$, i.e.\ no strict subset of $I$ is a model of $\PP$~\cite{ASPGel88}.
The \textit{reduct}~$\PP^I$ of a disjunctive ASP program $\PP$ w.r.t.\ an interpretation $I$ is defined as the simple disjunctive ASP program $\PP^I=\{a_1 \vee \hdots \vee a_k \la b_1,\hdots,b_m \,|\, (a_1 \vee \hdots \vee a_k \la b_1,\hdots,b_m,not \,c_1,\hdots, not \, c_n) \in \PP, \{c_1,\hdots,c_n\} \cap I = \es \}$. An interpretation $I$ of a disjunctive ASP program $\PP$ is an answer set of $\PP$ iff $I$ is an answer set of $\PP^I$.

\begin{example}
Let $\PP$ be the ASP program with the following 4 rules:
\begin{align*}
man(john) &\la,\quad person(john)\la,\quad person(fiona)\la \\
woman(X) \vee child(X) &\la person(X), not\, man(X)
\end{align*}
The first 3 rules are facts; hence their heads will be in any answer set. The fourth rule encodes that any person who is not a man, is a woman or child. The latter rule is grounded to 2 rules in which $X$ is resp.\ replaced by $john$ and $fiona$. We check that the interpretation $I=\{man(john), woman(fiona)$, $person(john)$, $person(fiona)\}$ is an answer set of the ground version of $\PP$ by computing the reduct~$\PP^I$. As the grounded rule with $X=john$ is deleted since $man(john)$ is in $I$, $\PP^I$ is:
\allowdisplaybreaks
\begin{align*}
man(john) &\la,\quad person(john)\la, \quad person(fiona)\la \\
woman(fiona) \vee child(fiona) &\la person(fiona)
\end{align*}
It is clear that $I$ is a minimal model of this simple program, so $I$ is an answer set of~$\PP$. By replacing $woman(fiona)$ by $child(fiona)$ in $I$, another answer set is obtained.
\end{example}
To automatically compute the answer sets of the programs in this paper, we have used the ASP solver DLV (\url{www.dlvsystem.com}), due to its ability to handle predicates, disjunction and numeric values, with built-in aggregate functions~\cite{ASPFa08}. The numeric values are only used for grounding.

\section{Modeling the Stable Marriage Problem in ASP} \label{sec:SMPinASP}
\subsection{Modeling the SMTI in ASP} \label{sec:smpinasp}
\nwf{In this section we model the SMTI, using ASP. A~few proposals of using non-monotonic reasoning for modeling the SMP have already been described in the literature.} For instance, in~\cite{ASPMa90} a specific variant of the SMP is mentioned (in which boys each know a subset of a set of girls and want to be matched to a girl they know) and in~\cite{SMPDung95} an abductive program is used to find a stable matching of marriages in which two fixed persons are paired, with strict, complete preference lists. To the best of our knowledge, beyond a few specific examples, no comprehensive study has been made of using ASP or related paradigms in this context. In particular, the generality of our ASP framework for weakly stable matchings of SMTI instances allows to easily adjust the encoding to variants of the SMP, such as the stable roommate problem~\cite{SMPGale62}\nw{, in which matches need to be found within one group instead of between two groups,} or the three-dimensional stable matching problem~\cite{SMPNg91}\nw{, which matches triples between three groups instead of pairs between two.}

The expression $accept(m,w)$ denotes that a man $m$ and a woman $w$ accept each other as partners. The predicate $manpropose(m,w)$ expresses that man $m$ is willing to propose to woman $w$ and analogously $womanpropose(m,w)$ expresses that woman $w$ is willing to propose to man $m$. Inspired by the Gale-Shapley algorithm, we look for an ASP formalization to find the stable matchings.
\begin{definition}[ASP program induced by SMTI] \label{def:aspsmti}
The ASP program $\PP$ induced by an instance $(\{\sigma_M^1,\hdots,\sigma_M^n\},\{\sigma_W^1,\hdots,\sigma_W^p\})$ of the SMTI is the program containing for every $i\in\{1,\hdots,n\},j\in\{1,\hdots,p\}$ the following rules:
\begin{align}
accept(m_i,w_j) &\la manpropose(m_i,w_j), womanpropose(m_i,w_j) \label{eq:ruleacc}\\
accept(m_i,m_i) &\la \{not\, accept(m_i,w_k) \,|\, k \in acceptable_M^i\} \label{eq:msingleindif} \\
accept(w_j,w_j) &\la \{not\, accept(m_k,w_j) \,|\, k \in acceptable_W^j\} \label{eq:wsingleindif}
\end{align}
and for every $i\in\{1,\hdots,n\}$, $j\in acceptable_M^i$:
\begin{align}
manpropose(m_i,w_j) &\la \{not\, accept(m_i,x)\,|\, x \leq_M^{m_i} w_j  \mbox{ and } w_j \neq x\} \label{eq:mpropindif}
\end{align}
and for every $j\in\{1,\hdots,p\}$, $i\in acceptable_W^j$:
\begin{align}
womanpropose(m_i,w_j) &\la \{not\, accept(x,w_j)\,|\, x \leq_W^{w_j} m_i  \mbox{ and } m_i \neq x\} \label{eq:wpropindif}
\end{align}
\end{definition}
Intuitively (\ref{eq:ruleacc}) means that a man and woman accept each other as partners if they propose to each other. Due to (\ref{eq:msingleindif}), a man accepts himself as a partner (i.e.\ stays single) if no woman in his preference list is prepared to propose to him. Rule (\ref{eq:mpropindif}) states that a man proposes to a woman if he is not paired to a more or equally preferred woman. For $j \in neutral_M^i$ the body of (\ref{eq:mpropindif}) contains $not \, accept(m_i,m_i)$. 
No explicit rules are stated about the number of persons someone can propose to or accept but as we will see below, in Proposition \ref{pr:SMPASP}, this is unnecessary.
\nw{Note that, for $k=\max(n,p)$, the number of grounded rules in the induced ASP program is~$\mathcal{O}(k^2)$.}
\nwf{We now illustrate our approach with an example.}
\begin{example} \label{ex:smpasp}
\nwf{Consider the following instance $(S_M,S_W)$ of the SMTI.
Let $M=\{m_1,m_2\}$ and $W=\{w_1,w_2,w_3\}$. 
Furthermore:}
\begin{align*}
\begin{aligned}
\sigma_M^1 &= (\{1\},\{2,3\},\{\})\\
\sigma_M^2 &= (\{2\},\{1\})\\
\end{aligned}
\hspace{30pt}
\begin{aligned}
\sigma_W^1 &= (\{1,2\},\{\})\\
\sigma_W^2 &= (\{1\},\{\})\\
\sigma_W^3 &= (\{2\},\{1\},\{\})
\end{aligned}
\end{align*}
\nwf{The ASP program induced by this SMTI instance is:}
\allowdisplaybreaks
\begin{align*}
accept(X,Y) &\la manpropose(X,Y),womanpropose(X,Y)\\
manpropose(m_1,w_1) &\la \\
manpropose(m_1,w_2) &\la not\, accept(m_1,w_1),not\, accept(m_1,w_3)\\
manpropose(m_1,w_3) &\la not\, accept(m_1,w_1),not\, accept(m_1,w_2)\\
accept(m_1,m_1) &\la not\, accept(m_1,w_1),not\, accept(m_1,w_2),\\
&\hspace{17pt}  not\, accept(m_1,w_3)\\
manpropose(m_2,w_2) &\la \\
manpropose(m_2,w_1) &\la  not\,accept(m_2,w_2),not\,accept(m_2,m_2)\\ 
accept(m_2,m_2) &\la not\,accept(m_2,w_2),not\,accept(m_2,w_1)\\
womanpropose(m_1,w_1) &\la not\,accept(m_2,w_1)\\
womanpropose(m_2,w_1) &\la not\,accept(m_1,w_1)\\ 
accept(w_1,w_1)  &\la not\,accept(m_1,w_1), not\,accept(m_2,w_1)\\
womanpropose(m_1,w_2) &\la\\
accept(w_2,w_2)  &\la not\,accept(m_1,w_2)\\
womanpropose(m_2,w_3) &\la \\
womanpropose(m_1,w_3) &\la not\, accept(m_2,w_3)\\
accept(w_3,w_3)  &\la not\,accept(m_1,w_3), not\,accept(m_2,w_3)
\end{align*}
\nwf{If we run the program in DLV, we get three answer sets, containing respectively:}
\begin{itemize}
\item $\{accept(m_1,w_3), accept(m_2,w_1), accept(w_2,w_2)\}$,
\item $\{accept(m_1,w_2), accept(m_2,w_1), accept(w_3,w_3)\}$,
\item $\{accept(m_1,w_1), accept(m_2,m_2), accept(w_2,w_2), accept(w_3,w_3)\}$.
\end{itemize}
\nwf{These answer sets correspond to the three weakly stable matching of marriages of this SMTI instance, namely $\{(m_1,w_3)$, $(m_2,w_1)$, $(w_2,w_2)\}$, $\{(m_1,w_2)$, $(m_2,w_1)$, $(w_3,w_3)\}$ and $\{(m_1,w_1)$, $(m_2,m_2)$, $(w_2,w_2),(w_3,w_3)\}$.}
\end{example}

\nwf{The following proposition states that our ASP encoding is sound, i.e.\ that there is a bijective correspondence between the answer sets of the induced program and the weak stable matchings of the SMTI. The complete proof is provided in the online appendix.}
\begin{proposition} \label{pr:SMPASP}
Let $(S_M,S_W)$ be an instance of the SMTI and let $\PP$ be the corresponding ASP program. If $I$ is an answer set of $\PP$, then a weakly stable matching for $(S_M,S_W)$ is given by $\{(x,y) \,|\, accept(x,y)\in I\}$. Conversely, if $\{(x_{1},y_{1})$, $\hdots$, $(x_{k},y_{k})\}$ is a weakly stable matching for $(S_M,S_W)$ then~$\PP$ has the following answer set~$I$:
\allowdisplaybreaks
\begin{align*}
&\{manpropose(x_{i},y)\,|\, i \in \{1,\hdots,k\}, x_{i}\in M, y <_M^{x_i} y_{i} \vee y = y_i \neq x_i\} \\
\cup &\{womanpropose(x,y_{i})\,|\,i \in \{1,\hdots,k\}, y_{i}\in W, x <_W^{y_i} x_i \vee x = x_i \neq y_i\} \\
\cup &\{accept(x_{i},y_{i}) \,|\, i \in \{1,\hdots,k\}\} 
\end{align*} 
\end{proposition}

A pair $(m,w)$ is \textit{stable} if there exists a stable matching that contains $(m,w)$. In~\cite{SMPMan02} it is shown that the decision problem `is the pair $(m,w)$ stable?' for a given SMTI instance is an NP-complete problem, even in the absence of unacceptability. It is straightforward to see that we can reformulate this decision problem as `does there exist an answer set of the induced normal ASP program~$\PP$ which contains the literal $accept(m,w)$?' (i.e.\ brave reasoning), which is known to be an NP-complete problem~\cite{ASPBa03}. Thus our model forms a suitable framework for these kind of decision problems concerning the SMTI.

\subsection{Modeling the 3-Dimensional SMTI in ASP}\label{sec:3dsmpinasp}
\nwf{
To illustrate further the flexibility of our ASP approach, we consider a variant of the SMP and show how small adaptations of the ASP encoding can solve this variant. Extending the SMP by adding another dimension to the problem was first proposed in~\cite{SMPKnuth76}.
 We work out the three-dimensional SMTI, where $n$ men are to be matched with $p$ women and $r$ children.
Definition~\ref{def:smti} can straightforwardly be generalized to a three-dimensional instance $(S_M,S_W,S_C)$, in which preference lists of the men are defined over the set of woman-child pairs and similarly the women have preferences over man-child pairs and the children have preferences over man-woman pairs.
In the three-dimensional case, $\sigma^i_M$ becoms a list of disjoint subsets of $\{1,\hdots,p\} \times \{1,\hdots,r\}$, and analogously for $\sigma^j_W$ and $\sigma^k_C$. Similarly as before, we can define the notions of $acceptable^i_M$ etc. Note that the orderings $\leq_M^{m_i}$, $\leq_W^{w_j}$ and $\leq_C^{c_k}$ are resp.\ defined on pairs in $W \times C$, $M\times C$ and $M\times W$.
A stable matching is now defined as a set of man-woman-child triples and singles, with the properties that no man, woman and child can be found such that each of them prefers the pair formed by the others above their current mates in the matching and no person prefers being single to being matched with its current mates. The practical relevance of this problem is pointed out in~\cite{SMPBi10}.}

\nwf{
Extending the ASP program from Definition~\ref{def:aspsmti}, we can write an ASP program induced by an instance of the three-dimensional matching problem.}
\begin{definition}[ASP program induced by 3D SMTI] \label{def:aspsmti3d}
The ASP program $\PP$ induced by an instance $(\{\sigma_M^1,\hdots,\sigma_M^n\},\{\sigma_W^1,\hdots,\sigma_W^p\},$ $\{\sigma_C^1,$ $\hdots,$ $\sigma_C^r\})$ of the 3-dimensional SMTI is the program containing the following rules for every $i\in\{1,\hdots,n\},j\in\{1,\hdots,p\}$ and $k\in\{1,\hdots,r\}$:
\begin{align*}
accept(m_i,w_j,c_k) &\la manprop(m_i,w_j,c_k), womprop(m_i,w_j,c_k),\\
&\hspace{17pt} childprop(m_i,w_j,c_k) \\
accept(m_i,m_i,m_i) &\la \{not\, accept(m_i,w_u,c_v) \,|\, (u,v) \in acceptable_M^i\} \\
accept(w_j,w_j,w_j) &\la \{not\, accept(m_u,w_j,c_v) \,|\, (u,v) \in acceptable_W^j\} \\
accept(c_k,c_k,c_k) &\la \{not\, accept(m_u,w_v,c_k) \,|\, (u,v) \in acceptable_C^k\} 
\end{align*}
and for every $i\in\{1,\hdots,n\}$ and $(j,k) \in acceptable_M^i$:
\begin{align*}
manprop(m_i,w_j,c_k) &\la \{not\, accept(m_i,x,y)\,|\, (x,y) \leq_M^{m_i} (w_j,c_k) ; (w_j,c_k) \neq (x,y)\} 
\end{align*}
and for every $j\in\{1,\hdots,p\}$ and $(i,k)\in acceptable_W^j$:
\begin{align*}
womprop(m_i,w_j,c_k) &\la \{not\, accept(x,w_j,y)\,|\, (x,y) \leq_W^{w_j} (m_i,c_k) ; (m_i,c_k) \neq (x,y)\} 
\end{align*}
and for every $k\in\{1,\hdots,r\}$ and $(i,j)\in acceptable_C^j$:
\begin{align*}
childprop(m_i,w_j,c_k) &\la \{not\, accept(x,u,c_k)\,|\, (x,y) \leq_C^{c_k} (m_i,w_j) ; (m_i,w_j) \neq (x,y)\} 
\end{align*}
\end{definition}
\nwf{
Ng and Hirschberg~\cite{SMPNg91} proved that deciding whether a stable matching exists for the three-dimensional problem -- in the absence of unacceptability and ties -- is an NP-complete problem.
Completely analogously as for the two-sided SMP, one can prove that this encoding yields a bijective correspondence between the answer sets of the ASP program and the stable matchings of the three-dimensional matching problem.
}
\nw{Note that, for $k=\max(n,p,r)$, the number of grounded rules in the induced ASP program is~$\mathcal{O}(k^3)$.}


\section{Selecting Optimal Stable Matchings} \label{sec:oss}

\subsection{Notions of Optimality of Stable Matchings}
When several stable matchings can be found for an instance of the SMP, some may be more interesting than others. The stable matching found by the G-S algorithm is \textit{M-optimal}~\cite{SMPRoth90}, i.e.\ every man likes this set at least as well as any other stable matching. 
Exchanging the roles of men and women in the G-S algorithm yields a \textit{W-optimal} stable matching~\cite{SMPGale62}, optimal from the women's point of view. 

While some applications may require us to favour either the men or the women, in others it makes more sense to treat both parties equally. To formalize some commonly considered notions of fairness and optimality w.r.t.\ the SMP, we define the cost $c_x(y)$ of being matched with $y$ for an individual $x$, where $c_x(y)=k$ if $y$ is $x$'s $k^{th}$ preferred partner. More precisely, for a man~$m_i$, we define $c_{m_i}(y) = | \{ z : z <_M^{m_i} y \} |+1$ for every $y \in acceptable_M^i$; for a woman~$w_j$, $c_{w_j}$ is defined analogously. So in case of ties we assign the same list position to equally preferred partners, as illustrated in Example \ref{ex:cost}.
\begin{example}\label{ex:cost}
Let $x=m_1$ be a man with preference list $\sigma_M^1 = (\{1\},\{2, 3\},\{4\})$ then $c_x(w_1)=1$, $c_x(w_2)=c_x(w_3)=2$ and $w_4$ yields $c_x(w_4)=4$. The cost for being single would be $4$, i.e.\ $c_x(m_1)=4$, since $m_1$ prefers women $w_1,w_2$ and $w_3$ to being single, but is indifferent between being paired to $w_4$ or staying single.
\end{example}
\begin{definition}[Optimal Stable Matchings]\label{def:optss}
Let $S$ be a set of marriages and let $S(x)$ denote the partner of $x$ in $S$.
\begin{itemize}
\item The sex-equality cost of $S$ is $c_{sexeq}(S)=|\sum_{x \in M}{c_x(S(x))} - \sum_{x \in W}{c_x(S(x))}|$,
\item the egalitarian cost of $S$ is $c_{weight}(S)=\sum_{x \in M \cup W}{c_x(S(x))}$, 
\item the regret cost of $S$ is $c_{regret}(S)=\max_{x \in M \cup W}{c_x(S(x))}$, and
\item the cardinality cost of $S$ is $c_{singles}(S)=|\{z : (z,z) \in S\}|$.
\end{itemize}
$S$ is a sex-equal stable matching iff $S$ is a stable matching with minimal sex-equality cost. Similarly, $S$ is an egalitarian (resp.\ minimum regret, maximum cardinality) stable matching iff $S$ is a stable matching with minimal egalitarian (resp.\ regret or cardinality) cost. 
\end{definition}
A \textit{sex-equal stable matching} assigns an equal importance to the preferences of the men and women, i.e.\ the men are as pleased with the matching as the women. An \textit{egalitarian stable matching} is a stable matching in which the preferences of every individual are considered to be equally important, i.e.\ it minimizes the difference in happiness of all the men and women. In~\cite{SMPXu11} the use of an egalitarian stable matching is proposed to optimally match virtual machines (VM) to servers in order to improve cloud computing by equalizing the importance of migration overhead in the data center network and VM migration performance.
A \textit{minimum regret stable matching} is optimal for the person who is worst off, i.e.\ there does not exist a stable matching such that the person who is most displeased with the matching is happier than the most displeased person in the minimum regret stable matching. A \textit{maximal or minimal cardinality stable matching} is a stable matching with resp.\ as few or as many singles as possible. Examples of practical applications include an efficient kidney exchange program~\cite{SMPRoth05}\nw{, which matches donors of incompatible pairs to recipients of other incompatible pairs and vice versa,} and the National Resident Matching Program (www.nrmp.org)~\cite{SMPMan02}\nw{, which matches healthcare professionals to graduate medical education and advanced training programs}. Maximizing cardinality guarantees that as many recipients as possible will get a compatible donor and as many healthcare professionals as possible will get a position.

\begin{remark}\label{note:termen}
\nwf{It might be somewhat confusing that the term \textit{utilitarian} is more frequently used in sociological and economical contexts for an optimization of the overall happiness (here called egalitarian), while \textit{egalitarian} is more used for an optimization which minimizes the unhappiness of the individuals (tending more to minimum regret). However, we will use the cost terms as defined above, since these are standard in the context of the SMP.}
\end{remark}

\nw{For an overview of literature results concerning the computational complexity of finding optimal stable matchings in the SMP, SMI, SMT and SMTI, we refer to the online appendix.}

\nw{Note that other notions of preferred matchings have been described in the literature, such as \textit{popularity}~\cite{SMPGard75}, \textit{Pareto optimality}~\cite{SMPGale85,SMPRoth90} or profile-based notions such as rank-maximum~\cite{SMPIrv03}, greedy maximum and generous maximum~\cite{SMPIrv03}. For more details on these and other optimality criteria, we refer the interested reader to~\cite{SMPMan13} for an overview.}

\subsection{Finding Stable Matchings using Disjunctive Naf-free ASP}
As we discuss in Section~\ref{sec:defOSSASP}, we can extend our ASP encoding of the SMTI such that the optimal stable matchings correspond to the answer sets of an associated ASP program. In particular, we use the saturation technique~\cite{ASPEi95} to filter non-optimal answer sets. 
Intuitively, the idea is to create a program with 3 components: (i) a first part describing the solution candidates, (ii) a second part also describing the solution candidates since comparison of solutions requires multiple solution candidates within the same answer set whereas the first part in itself produces one solution per answer set, (iii) a third part comparing the solutions described in the first two parts and selecting the preferred solutions by saturation.
It is, however, known that the presence of negation-as-failure can cause problems when applying saturation. 
\nw{This is due to the fact that rules containing naf-literals can be altered in the reduct. To address this issue, we use saturation in combination with a disjunctive naf-free ASP program instead of the ASP program in Definition~\ref{def:aspsmti}. To this end, we use a SAT encoding~\cite{ASPJan04} of the ASP program in Definition~\ref{def:aspsmti} and define a disjunctive naf-free ASP program in Definition~\ref{def:disjaspsmti} which selects particular models of the SAT problem.}

\nwf{Note that our original normal program is absolutely tight, i.e.\ there is no infinite sequence $l_1,l_2,\hdots$ of literals such that for every $i$ there is a program rule for which $l_{i+1}$ is a positive body literal and $l_i$ is in the head~\cite{ASPEr03}. 
We use the completion to derive an ASP encoding for finding optimal stable matchings. 
The completion of a normal ASP program is a set of propositional formulas. For every atom $a$ with $a \la body_i$ ($i \in \{1,\hdots,k\}$) all the program rules with head $a$, the propositional formula $a \equiv body'_1 \vee \hdots \vee body'_k$ is in the completion of that program, 
\nw{where $body'_i$ is the conjunction of literals derived from $body_i$ by replacing every occurrence of `$not$' with `$\neg$'.}
If an atom $a$ of the program does not occur in any rule head, then $a \equiv \, \perp$ is in the completion of the program. Similarly the completion of the program contains the propositional formula $\perp \,\equiv body'_1 \vee \hdots \vee body'_l$ where the disjunction extends over all the program constraints $\la body_i$ ($i \in \{1,\hdots,l\}$). Because our program is absolutely tight, we know that every propositional model of the completion will correspond to an answer set of the original program and vice versa~\cite{ASPEr03}. 
When applied to the induced normal ASP program in Definition \ref{def:aspsmti}, the completion contains the following formulas, for all $i \in\{1,\hdots,n\}$ and $j\in\{1,\hdots,p\}$: }
\allowdisplaybreaks
\begin{align*}
&accept(m_i,w_j) \equiv manpropose(m_i,w_j) \wedge womanpropose(m_i,w_j),Ê\\
&accept(m_i,m_i) \equiv \bigwedge_{k \in acceptable_M^i}{ \neg accept(m_i,w_k)}, \\
&accept(w_j,w_j) \equiv \bigwedge_{k \in acceptable_W^j}{ \neg accept(m_k,w_j)},
\end{align*}
and for all $i \in\{1,\hdots,n\}$ and $j\in acceptable_M^i$:
\begin{align*}
manpropose(m_i,w_j) \equiv \bigwedge_{x \leq_M^{m_i} w_j, x \neq w_j}{\neg accept(m_i,x)} ,
\end{align*}
and similarly for all $j\in\{1,\hdots,p\}$ and $i \in acceptable_W^j$:
\begin{align*}
womanpropose(m_i,w_j) \equiv \bigwedge_{x \leq_W^{w_j} m_i, x \neq m_i}{\neg accept(x,w_j)} ,
\end{align*}
and for all $i \in \{1,\hdots,n\}$ and $j \in unacceptable_M^i$:
\begin{align*}
manpropose(m_i,w_j) \equiv \,\perp ,
\end{align*}
and similarly for all $j \in \{1,\hdots,p\}$ and $i \in unacceptable_W^j$:
\begin{align*}
womanpropose(m_i,w_j) \equiv \,\perp.
\end{align*}
\nwf{Using these formulas, which form the completion of the normal ASP program from Definition \ref{def:aspsmti}, we can define an equivalent disjunctive ASP program without negation-as-failure. 
\begin{definition}[Induced disj.\ naf-free ASP program] \label{def:disjaspsmti}
The disjunctive naf-free ASP program $\PP_{disj}$ induced by an SMTI instance $(S_M,S_W)$ contains the following rules for every $i\in\{1,\hdots,n\},j\in\{1,\hdots,p\}$:
\allowdisplaybreaks
\begin{align*}
\neg accept(m_i,w_j) \vee manpropose(m_i,w_j) &\la \\
\neg accept(m_i,w_j) \vee womanpropose(m_i,w_j) &\la\\
accept(m_i,w_j) \vee \neg manpropose(m_i,w_j) \vee \neg womanpropose(m_i,w_j) &\la
\end{align*}
For every $i\in\{1,\hdots,n\}$, $l\in unacceptable_M^i$, $j \in acceptable_M^i$, $x \leq_M^{m_i} w_j, x\neq w_j$ $\PP_{disj}$  contains:
\begin{align*}
\bigvee_{k \in acceptable_M^i} accept(m_i,w_k) \vee accept(m_i,m_i) &\la\\
\neg accept(m_i,m_i) \vee \neg accept(m_i,w_j) &\la \\
\neg manpropose(m_i,w_j) \vee \neg accept(m_i,x) &\la \\
\bigvee_{x \leq_M^{m_i} w_j, x \neq w_j} accept(m_i,x) \vee manpropose(m_i,w_j) &\la\\
\neg manpropose(m_i,w_l) &\la
\end{align*}
and symmetrical for $j \in \{1,\hdots,p\}$ and $womanpropose$.
\end{definition}
\nw{Note that, for $k=\max(n,p)$, the number of grounded rules in the induced naf-free program is~$\mathcal{O}(k^3)$.}
The following lemma follows from the fact that the completion corresponds to the original program~\cite{ASPEr03}.}
\begin{lemma} \label{lem:disj}
Let $\PP$ be the normal ASP program from Definition \ref{def:aspsmti} and $\PP_{disj}$ the disjunctive ASP program from Definition \ref{def:disjaspsmti}. It holds that for any answer set $I$ of $\PP$ there exists an answer set $I_{disj}$ of $\PP_{disj}$ such that the atoms of $I$ and $I_{disj}$ coincide. Conversely for any answer set $I_{disj}$ of $\PP_{disj}$ there exists an answer set $I$ of $\PP$ such that the atoms of $I$ and $I_{disj}$ coincide.
\end{lemma}

\subsection{ASP Program to Select Optimal Solutions} \label{sec:defOSSASP}
Let $(S_M,S_W)$ be an SMTI instance with $S_M = \{\sigma_M^1,\hdots,\sigma_M^n\}$ and $S_W = \{\sigma_W^1,\hdots,$ $\sigma_W^p\}$, and let $\PP_{norm}$ be the induced normal ASP program from Definition \ref{def:aspsmti}. Our technique for extending this program to a program that can respectively optimize for the sex-equality, egalitarian, minimum regret and maximum cardinality criterion is in each case very similar. We start by explaining it for the case of sex-equality. Our first step is to add a set of rules that compute the sex-equality cost of a~set of~marriages. For every man $m_i$ and every $x \in acceptable_M^i \cup \{m_i\}$ we use the following rule to determine the cost for $m_i$:
\begin{align}
mancost(i,c_{m_i}(x)) &\la accept(m_i,x) \label{eq:costm}
\end{align}
and similarly for every $w_j$ and every $x \in acceptable_W^j\cup \{w_j\}$:
\begin{align}
womancost(j,c_{w_j}(x)) &\la accept(x,w_j) \label{eq:costw}
\end{align}
We also use the following rules with $i$ ranging from $1$ to $n$ and $j$ from $1$ to $p$:
\allowdisplaybreaks
\begin{align}
manweight(Z) &\la \#sum\{B,A : mancost(A,B)\}=Z, \#int(Z) \label{eq:weightm}\\
womanweight(Z) &\la \#sum\{B,A : womancost(A,B)\}=Z, \#int(Z)  \label{eq:weightw}\\
sexeq(Z) \la man&weight(X), womanweight(Y), Z=X-Y \nonumber\\
sexeq(Z) \la man&weight(X), womanweight(Y), Z=Y-X \label{eq:sexeq}
\end{align}
Note that $\#sum$, $\#max$, $\#int$ and $\#count$ are DLV aggregate functions~\cite{ASPFa08}. The `$A$' mentioned as variable in $\#sum$ indicates that a cost must be included for every person (otherwise the cost is included only once when persons have the same cost).
Rule (\ref{eq:weightm}) determines the sum of the male costs and similarly (\ref{eq:weightw}) determines the sum of the female costs. According to Definition \ref{def:optss} the absolute difference of these values yields the sex-equality cost, as determined by rules (\ref{eq:sexeq}). Since numeric variables are restricted to positive integers in DLV, we omit conditions as `$X\geq Y$' or `$X < Y$'. The program $\PP_{norm}$ extended with rules (\ref{eq:costm}) -- (\ref{eq:sexeq}) is denoted $\PP^{sexeq}_{ext}$.

We construct a program $\PP_{sexeq}$, composed by subprograms, that selects optimal solutions. Let $\PP'_{disj}$ be the disjunctive naf-free ASP program, induced by the same SMTI instance, in which a prime symbol is added to all literal names (e.g.\ $accept$ becomes $accept'$). Define a new program $\PP'^{sexeq}_{ext}$ with all the rules of $\PP'_{disj}$ in which every occurrence of $\neg atom$ is changed into $natom$ for every atom $atom$, i.e.\ replace all negation symbols by a prefix `$n$'. For every occurring atom $atom$ in $\PP'^{sexeq}_{ext}$, add the following rule to exclude non-consistent solutions:
\begin{align}
sat &\la atom, natom \label{eq:geenneg}
\end{align}
For instance, the rule $sat \la accept'(m_1,w_1), naccept'(m_1,w_1)$ is added.
Finally add rules (\ref{eq:costm}) -- (\ref{eq:sexeq}) with prime symbols to the literal names to $\PP'^{sexeq}_{ext}$ but replace rule (\ref{eq:weightm}) and rule (\ref{eq:weightw}) by:
\begin{align}
mansum'(n,X) &\la mancost'(n,X) \nonumber\\
mansum'(J,Z) &\la mansum'(I,X), mancost'(J,Y), Z=X+Y, \#succ(J,I) \nonumber\\ 
manweight'(Z) &\la mansum'(1,Z) \nonumber \\
womansum'(p,X) &\la womancost'(p,X) \nonumber\\
womansum'(J,Z) &\la womansum'(I,X), womancost'(J,Y), Z=X+Y, \nonumber\\
&\hspace{13pt} \#succ(J,I) \nonumber\\ 
womanweight'(Z) &\la womansum'(1,Z) 
\label{eq:weightsucc}
\end{align}
\nwf{The DLV aggregate function $\#succ(J,I)$ is true whenever $J+1=I$. We replace the rules with the aggregate function $\#max$ by these rules to make sure the saturation happens correctly. When saturation is used, the DLV aggregate functions $\#max$, $\#sum$ and $\#count$ would not yield the right criterion values. Moreover, DLV does not accept these aggregate functions in saturation because of the cyclic dependency of literals within the aggregate functions created by the rules for saturation. These adjusted rules, however, will not pose any problems because of the successive way they compute the criterion values. This becomes more clear in the proof of Proposition \ref{pr:optssASP}.} We define the ASP program $\PP_{sexeq}$ as the union of $\PP^{sexeq}_{ext}$, $\PP'^{sexeq}_{ext}$ and $\PP_{sat}$. The ASP program $\PP_{sat}$ contains the following rules to select minimal solutions based on sex-equality:
\allowdisplaybreaks
\begin{align} \label{eq:satcrit}
sat  &\la sexeq(X), sexeq'(Y), X \leq Y \\
&\la not \,  sat \label{eq:sat}\\
mancost'(X,Y) &\la  sat, manargcost'_1(X), manargcost'_2(Y) \nonumber \\
womancost'(X,Y) &\la  sat, womanargcost'_1(X), womanargcost'_2(Y)  \label{eq:cost}\\
manpropose'(X,Y) &\la  sat, man(X), woman(Y)  \nonumber\\
womanpropose'(X,Y) &\la  sat, man(X), woman(Y) \nonumber \\
accept'(X,X) & \la  sat, man(X) \nonumber\\
accept'(X,X) & \la  sat, woman(X) \nonumber \\
accept'(X,Y) &\la  sat, man(X), woman(Y) \label{eq:satur}
\end{align}
and analogous to (\ref{eq:satur}) a set of rules with prefix `$n$' for the head predicates.
Finally we add the facts $manargcost'_1(1..n) \la$, $manargcost'_2(1..(p+1)) \la$, $womanargcost'_1(1..p) \la$, $womanargcost'_2(1..(n+1)) \la$, $man(x) \la$ for every man $x$ and $woman(x) \la$ for every woman $x$ to $\PP_{sat}$.
The rule $manargcost'_1(1..n) \la$ is DLV syntax for the $n$ facts $manargcost'_1(1) \la, \hdots, manargcost'_1(n) \la$.
Intuitively the rules of $\PP_{sat}$ express the key idea of saturation. First every answer set is forced to contain the atom $sat$ by rule (\ref{eq:sat}). Then the rules (\ref{eq:cost}) -- (\ref{eq:satur}) and the facts make sure that any answer set should contain all possible literals with a prime symbol that occur in $\PP_{sexeq}$. Rule (\ref{eq:satcrit}) will establish that only optimal solutions will correspond to minimal models and thus lead to answer sets. For any non-optimal solution, the corresponding interpretation containing $sat$ will never be a minimal model of the reduct. It is formally proven in Proposition \ref{pr:optssASP} below that $\PP_{sexeq}$ produces exactly the stable matchings with minimal sex-equality cost.

Furthermore, only small adjustments to $\PP_{sexeq}$ are needed to create programs $\PP_{weight}$, $\PP_{regret}$, and $\PP_{singles}$ that respectively produce egalitarian, minimum regret and maximum cardinality stable matchings. Indeed, the ASP program $\PP_{weight}$ can easily be defined as $\PP_{sexeq}$ in which the predicates $sexeq$ and $sexeq'$ are respectively replaced by $weight$ and $weight'$ and the rules (\ref{eq:sexeq}) are replaced by (\ref{eq:weight}), determining the egalitarian cost of Definition \ref{def:optss} as the sum of the male and female costs:
\begin{align}
weight(Z) &\la manweight(X),womanweight(Y),Z=X+Y\label{eq:weight}
\end{align}

Similarly the ASP program $\PP_{regret}$ is defined as $\PP_{sexeq}$ in which the predicates $sexeq$ and $sexeq'$ are resp.\ replaced by $regret$ and $regret'$ and rules (\ref{eq:weightm}) -- (\ref{eq:sexeq}) are replaced by the following rules:
\begin{align}
manregret(Z) &\la \#max\{B : mancost(A,B)\}=Z, \#int(Z) \label{eq:regretm}\\
womanregret(Z) &\la \#max\{B : womancost(A,B)\}=Z, \#int(Z) \label{eq:regretw}\\
regret(X) &\la manregret(X), womanregret(Y), X>Y \nonumber\\
regret(Y) &\la manregret(X), womanregret(Y), X \leq Y \label{eq:regret}
\end{align} 
Rule (\ref{eq:regretm}) determines the regret cost but only for the men. Similarly (\ref{eq:regretw}) determines the regret cost for the women. The regret cost as defined in Definition \ref{def:optss} is the maximum of these two values, determined by the rules in (\ref{eq:regret}).
\nwf{Again we adjust rules (\ref{eq:regretm}) and (\ref{eq:regretw}) for the program part $\PP'^{regret}_{ext}$ by replacing them with a variant based on the successor function:}
\begin{align}
manmax'(n,X) &\la mancost'(n,X)\nonumber\\
manmax'(J,X) &\la manmax'(I,X), mancost'(J,Y), X \geq Y, \#succ(J,I)\nonumber\\
manmax'(J,Y) &\la manmax'(I,X), mancost'(J,Y), X < Y, \#succ(J,I)\nonumber\\
manregret'(Z) &\la manmax'(1,Z) \nonumber\\
womanmax'(p,X) &\la womancost'(p,X)\nonumber\\
womanmax'(J,X) &\la womanmax'(I,X), womancost'(J,Y), X \geq Y, \#succ(J,I)\nonumber\\
womanmax'(J,Y) &\la womanmax'(I,X), womancost'(J,Y), X < Y, \#succ(J,I)\nonumber\\
womanregret'(Z) &\la womanmax'(1,Z) 
\label{eq:regretsucc}
\end{align} 

Finally we define the ASP program $\PP_{singles}$ as $\PP_{sexeq}$ in which the predicates $sexeq$ and $sexeq'$ are resp.\ replaced by $singles$ and $singles'$. \nwf{Furthermore we replace rules (\ref{eq:costm}) -- (\ref{eq:sexeq}) by (\ref{eq:card}), determining the number of singles:}
\begin{align}
singles(Z) \la \#count\{B : accept(B,B)\} = Z, \#int(Z) \label{eq:card}
\end{align}
\nwf{This time we adjust rule (\ref{eq:card}) for the program part $\PP'^{singles}_{ext}$ as follows:}
\begin{align}
single'(p+i,1) &\la accept'(m_i,m_i), \hspace{14pt} single'(p+i,0) \la naccept'(m_i,m_i)\nonumber\\
single'(j,1) &\la accept'(w_j,w_j), \hspace{35pt} single'(j,0) \la naccept'(w_j,w_j)\nonumber\\
singlesum'(n+p,X) &\la single'(n+p,X)\nonumber\\
singlesum'(J,Z) &\la singlesum'(I,X), single'(J,Y), Z=X+Y, \#succ(J,I)\nonumber\\
singles'(Z) &\la singlesum'(1,Z) \label{eq:singlessucc}
\end{align}
\nw{Note that, for $k=\max(n,p)$, the number of grounded rules in the induced ASP program is~$\mathcal{O}(k^3)$ for minimum regret and maximum cardinality, but~$\mathcal{O}(k^4)$ for sex-equalness and egalitarity. The latter programs have a higher number of grounded rules because of how the weights are counted in the first and second program part.}
\nwf{We illustrate our method with an example.}
\begin{example}
\nwf{We reconsider Example \ref{ex:smpasp}. This SMTI instance had 3 stable matchings of marriages:}
\begin{itemize}
\item $S_1 = \{accept(m_1,w_3), accept(m_2,w_1), accept(w_2,w_2)\}$,
\item $S_2 = \{accept(m_1,w_2), accept(m_2,w_1), accept(w_3,w_3)\}$,
\item $S_3 = \{accept(m_1,w_1), accept(m_2,m_2), accept(w_2,w_2), accept(w_3,w_3)\}$.
\end{itemize}
\nwf{It is easy to compute the respective regret costs as $c_{regret}(S_1) = 2$ and $c_{regret}(S_2) = c_{regret}(S_3) = 3$. The corresponding ASP program selecting this minimum regret stable matching is the program consisting of the rules in Example~\ref{ex:smpasp} in addition to:}
\allowdisplaybreaks
\begin{align*}
man(m_1) &\la, \quad man(m_2) \la, \quad\\
woman(w_1) &\la, \quad woman(w_2) \la, \quad woman(w_3) \la\\
mancost(1,1) &\la accept(m_1,w_1), \hspace{13pt} womancost(1,1) \la accept(m_1,w_1)\\
mancost(1,2) &\la accept(m_1,w_2), \hspace{13pt} womancost(1,1) \la accept(m_2,w_1)\\
mancost(1,2) &\la accept(m_1,w_3), \hspace{13pt} womancost(1,2) \la accept(w_1,w_1)\\
mancost(1,4) &\la accept(m_1,m_1), \hspace{12pt} womancost(2,1) \la accept(m_1,w_2)\\
mancost(2,2) &\la accept(m_2,w_1), \hspace{13pt} womancost(2,2) \la accept(w_2,w_2)\\
mancost(2,1) &\la accept(m_2,w_2), \hspace{13pt} womancost(3,2) \la accept(m_1,w_3)\\
mancost(2,2) &\la accept(m_2,m_2), \hspace{12pt} womancost(3,1) \la accept(m_2,w_3)\\
& \hspace{101pt} womancost(3,3) \la accept(w_3,w_3)\\
manregret(Z) &\la \#max\{B: mancost(A,B)\}=Z, \#int(Z)\\
womanregret(Z) &\la \#max\{B: womancost(A,B)\}=Z, \#int(Z)\\
regret(X) &\la manregret(X), womanregret(Y), X>Y\\
regret(Y) &\la manregret(X), womanregret(Y), X<=Y
\end{align*}
\begin{align*}
naccept'(M,W) \vee manpropose'(M,W) &\la man(M), \\
&\hspace{8pt}woman(W)\\
naccept'(M,W) \vee womanpropose'(M,W) &\la man(M), \\
&\hspace{8pt}woman(W)\\
accept'(M,W) \vee nmanpropose'(M,W) \vee nwomanpropose'(M,W) &\la man(M), \\
&\hspace{8pt} woman(W)\\
accept'(m_1,w_1) \vee accept'(m_1,w_2) \vee accept'(m_1,w_3) \vee accept'(m_1,m_1) &\la \\
accept'(m_2,w_1) \vee accept'(m_2,w_2) \vee accept'(m_2,m_2) &\la \\
naccept'(m_1,m_1) \vee naccept'(m_1,w_1) &\la \\
naccept'(m_1,m_1) \vee naccept'(m_1,w_2) &\la \\
naccept'(m_1,m_1) \vee naccept'(m_1,w_3) &\la \\
naccept'(m_2,m_2) \vee naccept'(m_2,w_1) &\la \\
naccept'(m_2,m_2) \vee naccept'(m_2,w_2) &\la \\
accept'(m_1,w_1) \vee accept'(m_2,w_1) \vee accept'(w_1,w_1) &\la \\
accept'(m_1,w_2) \vee accept'(w_2,w_2) &\la \\
accept'(m_1,w_3) \vee accept'(m_2,w_3) \vee accept'(w_3,w_3) &\la \\
naccept'(w_1,w_1) \vee naccept'(m_1,w_1) &\la \\
naccept'(w_1,w_1) \vee naccept'(m_2,w_1) &\la \\
naccept'(w_2,w_2) \vee naccept'(m_1,w_2) &\la \\
naccept'(w_3,w_3) \vee naccept'(m_1,w_3) &\la \\
naccept'(w_3,w_3) \vee naccept'(m_2,w_3) &\la \\
nmanpropose'(m_1,w_2) \vee naccept'(m_1,w_1) &\la \\
nmanpropose'(m_1,w_2) \vee naccept'(m_1,w_3) &\la \\
nmanpropose'(m_1,w_3) \vee naccept'(m_1,w_1) &\la \\
nmanpropose'(m_1,w_3) \vee naccept'(m_1,w_2) &\la \\
manpropose'(m_1,w_1) &\la \\
accept'(m_1,w_1) \vee accept'(m_1,w_3) \vee manpropose'(m_1,w_2) &\la \\
accept'(m_1,w_1) \vee accept'(m_1,w_2) \vee manpropose'(m_1,w_3) &\la \\
nmanpropose'(m_2,w_1) \vee naccept'(m_2,w_2) &\la \\
nmanpropose'(m_2,w_1) \vee naccept'(m_2,m_2) &\la \\
manpropose'(m_2,w_2) &\la \\
accept'(m_2,w_2) \vee accept'(m_2,m_2) \vee manpropose'(m_2,w_1) &\la \\
nwomanpropose'(m_1,w_1) \vee naccept'(m_2,w_1) &\la \\
nwomanpropose'(m_2,w_1) \vee naccept'(m_1,w_1) &\la \\
accept'(m_1,w_1) \vee womanpropose'(m_2,w_1) &\la \\
accept'(m_2,w_1) \vee womanpropose'(m_1,w_1) &\la \\
womanpropose'(m_1,w_2) &\la \\
nwomanpropose'(m_1,w_3) \vee naccept'(m_2,w_3) &\la \\
womanpropose'(m_2,w_3) &\la \\
accept'(m_2,w_3) \vee womanpropose'(m_1,w_3) &\la \\
nmanpropose'(m_2,w_3) &\la\\
nwomanpropose'(m_2,w_2) &\la 
\end{align*}
\begin{align*}
sat &\la manpropose'(X,Y), nmanpropose'(X,Y), man(X), \\
&\hspace{17pt}woman(Y)\\
sat &\la womanpropose'(X,Y), nwomanpropose'(X,Y), \\
&\hspace{17pt}man(X), woman(Y)\\
sat &\la accept'(X,Y), naccept'(X,Y), man(X), woman(Y)\\
sat &\la accept'(X,X), naccept'(X,X), man(X)\\
sat &\la accept'(X,X), naccept'(X,X), woman(X)\\
mancost'(1,1) &\la accept'(m_1,w_1), \hspace{13pt} womancost'(1,1) \la accept'(m_1,w_1)\\
mancost'(1,2) &\la accept'(m_1,w_2), \hspace{13pt} womancost'(1,1) \la accept'(m_2,w_1)\\
mancost'(1,2) &\la accept'(m_1,w_3), \hspace{13pt} womancost'(1,2) \la accept'(w_1,w_1)\\
mancost'(1,4) &\la accept'(m_1,m_1), \hspace{12pt} womancost'(2,1) \la accept'(m_1,w_2)\\
mancost'(2,2) &\la accept'(m_2,w_1), \hspace{13pt} womancost'(2,2) \la accept'(w_2,w_2)\\
mancost'(2,1) &\la accept'(m_2,w_2), \hspace{13pt} womancost'(3,2) \la accept'(m_1,w_3)\\
mancost'(2,2) &\la accept'(m_2,m_2), \hspace{12pt} womancost'(3,1) \la accept'(m_2,w_3)\\
& \hspace{101pt} womancost'(3,3) \la accept'(w_3,w_3)\\
manmax'(2,X) &\la mancost'(2,X)\\
manmax'(J,X) &\la manmax'(I,X), mancost'(J,Y), X>=Y, \#succ(J,I)\\
manmax'(J,X) &\la manmax'(I,X), mancost'(J,Y), X>=Y, \#succ(J,I)\\
manregret'(Z) &\la manmax'(1,Z)\\
womanmax'(3,X) &\la womancost'(3,X)\\
womanmax'(J,X) &\la womanmax'(I,X), womancost'(J,Y), X>=Y, \\
&\hspace{14pt}\#succ(J,I)\\
womanmax'(J,X) &\la womanmax'(I,X), womancost'(J,Y), X>=Y, \\
&\hspace{14pt} \#succ(J,I)\\
womanregret'(Z) &\la womanmax'(1,Z)\\
regret'(X) &\la manregret'(X), womanregret'(Y), X>Y\\
regret'(Y) &\la manregret'(X), womanregret'(Y), X<=Y\\
sat &\la regret(X), regret'(Y), X<=Y\\
&\la not\, sat\\
manargcost_1'(1\mbox{..}2) &\la, \hspace{40pt} womanargcost_1'(1\mbox{..}3) \la \\
manargcost_2'(1\mbox{..}4) &\la, \hspace{40pt} womanargcost_2'(1\mbox{..}3) \la  \\
mancost'(X,Y) &\la sat, manargcost_1'(X),manargcost_2'(Y)\\
womancost'(X,Y) &\la sat, womanargcost_1'(X),womanargcost_2'(Y)\\
manpropose'(X,Y) &\la sat, man(X), woman(Y)\\
nmanpropose'(X,Y) &\la sat, man(X), woman(Y)\\
womanpropose'(X,Y) &\la sat, man(X), woman(Y)\\
nwomanpropose'(X,Y) &\la sat, man(X), woman(Y)\\
accept'(X,Y) &\la sat, man(X), woman(Y)\\
accept'(X,X) &\la sat, man(X)\\
accept'(X,X) &\la sat, woman(X)\\
naccept'(X,Y) &\la sat, man(X), woman(Y)\\
naccept'(X,X) &\la sat, man(X)\\
naccept'(X,X) &\la sat, woman(X)
\end{align*}
\nwf{Computing the unique answer set of this disjunctive ASP program with DLV and filtering it to the literals $accept$ and $regret$, yields $\{accept(m_2,w_1)$, $accept(m_1,w_3)$, $accept(w_2,w_2)$, $regret(2)\}$, corresponding exactly to the minimum regret stable matching $S_1$ of the SMTI instance and the corresponding regret cost.}
\end{example}

\nwf{We prove that there exists a bijective correspondence between the answer sets of the induced disjunctive ASP program and the optimal stable matchings of the SMTI (see the online appendix).}
\begin{proposition} \label{pr:optssASP}
Let the criterion $crit$ be an element of $\{sexeq$, $weight$, $regret$, $singles\}$. For every answer set $I$ of the program $\PP_{crit}$ induced by an SMTI instance the set $S_I = \{(m,w) \,|\,$ $accept(m$, $w)$ $\in$ $I\}$ forms an optimal stable matching of marriages w.r.t.\ criterion $crit$ and the optimal criterion value is given by the unique value $v_I$ for which $crit(v_I) \in I$. Conversely for every optimal stable matching $S = \{(x_1,y_1), \hdots,$ $(x_k,y_k)\}$ with optimal criterion value $v$ there exists an answer set $I$ of $\PP_{crit}$ such that $\{(x,y) \,|\,$ $accept(x,y)$ $\in I\} = \{(x_i,y_i) \,|\, i \in \{1,\hdots,k\}\}$ and $v$ is the unique value for which $crit(v) \in I$.
\end{proposition}

\begin{remark}
If we remove from $\PP_{sexeq}$ the rules (\ref{eq:weightw}) -- (\ref{eq:sexeq}) and replace rule (\ref{eq:satcrit}) by the rule $sat \la manweight(X)$, $manweight'(Y), X \leq Y$, then we obtain the M-optimal stable matchings. Analogously we can obtain the W-optimal stable matchings.

If a criterion is to be maximized, the symbol $\leq$ in rule (\ref{eq:satcrit}) is simply replaced by~$\geq$. E.g.\ for $crit=singles$ we will get minimum cardinality stable matchings.
\end{remark}

\section{Conclusion}
\nwf{We have shown how ASP programs can be used to encode a number of variations and generalizations of the SMP. Apart from the availability of efficient ASP solvers, the main advantage of our approach is its flexibility, allowing us to find solutions for a wide range of stable matching problems. We can, for instance, compute stable matchings of variants such as the three-dimensional stable matching problem, as well as select stable matchings based on optimality criteria, even for problems with unacceptable partners and ties. We have illustrated our method for sex-equality, egalitarity, minimum regret and maximum cardinality, but the approach can readily be adapted to other optimality criteria (e.g.\ popular matchings) or to different matching problems (e.g.\ the roommate problem). 
To the best of our knowledge, no other exact algorithms exist to find an optimal stable matching for an SMP instance with ties, regardless of the presence of unacceptability and regardless of whether the optimality notion is sex-equality, egalitarity, minimum regret or maximum cardinality.
Therefore, our encoding offers the first exact implementation for solving the aforementioned problems. 
 }

\bibliographystyle{acmtrans}
\bibliography{biblio}

\end{document}


\maketitle

\thispagestyle{myheadings}
\pagestyle{myheadings}

\label{firstpage}

\begin{appendix}
\section{Definition SMP and SMI}\label{sec:def}
\begin{definition}[SMP] 
An instance of the SMP is a pair $(S_M,S_W)$ with $S_M = \{\sigma_M^1,\hdots,\sigma_M^n\}$ and $S_W = \{\sigma_W^1,\hdots,\sigma_W^n\}$. For every $i\in\{1,\hdots,n\}$, $\sigma_M^i$ and $\sigma_W^i$ are permutations of $\{1,\hdots,n\}$. We call $\sigma_M^i$ and $\sigma_W^i$ the preferences of man $m_i$ and woman $w_i$ respectively. If $k = \sigma_M^i(j)$, woman $w_k$ is man $m_i$'s $j^{th}$ most preferred woman. The case $k = \sigma_W^i(j)$ is similar. Man $m$ and woman $w$ form a blocking pair in a set of marriages $S$ if $m$ strictly prefers $w$ to his partner in $S$ and $w$ strictly prefers $m$ to her partner in~$S$. A~weakly stable matching is a set of marriages without blocking pairs or individuals. 
\end{definition}
\begin{definition}[SMI] 
An instance of the SMI is a pair $(S_M,S_W)$ with $S_M = \{\sigma_M^1,\hdots,\sigma_M^n\}$ and $S_W = \{\sigma_W^1,\hdots,\sigma_W^p\}$. For every $i\in\{1,\hdots,n\}$, $\sigma_M^i$ is a permutation of a subset of $\{1,\hdots,p\}$. Symmetrically $\sigma_W^i$ is a permutation of a subset of $\{1,\hdots,n\}$ for every $i\in\{1,\hdots,p\}$. We call $\sigma_M^i$ and $\sigma_W^i$ the preferences of man $m_i$ and woman $w_i$ respectively. If $k = \sigma_M^i(j)$, woman $w_k$ is man $m_i$'s $j^{th}$ most preferred woman. The case $k \in \sigma_W^i(j)$ is similar. If there is no $l$ such that $j \in \sigma_M^i(l)$, woman $w_j$ is an unacceptable partner for man $m_i$, and similarly when there is no $l$ such that $j \in \sigma_W^i(l)$. Man $m$ and woman $w$ form a blocking pair in a set of marriages $S$ if $m$ strictly prefers $w$ to his partner in $S$ and $w$ strictly prefers $m$ to her partner in~$S$. A blocking individual in $S$ is a person who stricly prefers being single to being paired to his partner in $S$. A~weakly stable matching is a set of marriages without blocking pairs or individuals. 
\end{definition}

\section{Complexity results}\label{sec:table}
Table \ref{tab:compl} presents an overview of known complexity results concerning finding an optimal stable set\footnote{We assume that P $\neq$ NP}. 
\begin{table}[h]
\caption{Literature complexity results for finding an optimal stable set}
\begin{center}
\begin{tabular}{c|cc}
& sex-equal & egalitarian \\
\hline
SMP & NP-hard~\cite{SMPKat93} & P ($O(n^4)$~\cite{SMPIrv87}) \\
SMI & NP-hard~\cite{SMPMc12} & \\
SMT & & \\
SMTI & & \\
\end{tabular}
\end{center}
\begin{center}
\begin{tabular}{c|cc}
& min.\ regret & max.\ card.\ \\
\hline
SMP & P ($O(n^2)$~\cite{SMPGus87}) & P ($O(n^2)$~\cite{SMPGale62})\\
SMI & & P~\cite{SMPGale85}\\
SMT & NP-hard~\cite{SMPMan02} &\\
SMTI & & NP-hard~\cite{SMPMan99,SMPMan02}\\
\end{tabular}
\end{center}\label{tab:compl}
\end{table}
\vspace{-0.5cm}

\noindent
Between brackets we mention in Table \ref{tab:compl} the complexity of an algorithm that finds an optimal stable set if one exists, in function of the number of men $n$. To the best of our knowledge, the only exact algorithm tackling an NP-hard problem from Table \ref{tab:compl} finds a sex-equal stable set for an SMP instance in which the strict preference lists of men and/or women are bounded in length by a constant~\cite{SMPMc12}. To the best of our knowledge, no exact implementations exist to find an optimal stable set for an SMP instance with ties, regardless of the presence of unacceptability and regardless which notion of optimality from Table~\ref{tab:compl} is used. Our approach yields an exact implementation of all problems mentioned in Table~\ref{tab:compl}.

\section{Proof of Proposition 1}\label{sec:proof1}
\begin{proposition} 
Let $(S_M,S_W)$ be an instance of the SMTI and let $\PP$ be the corresponding ASP program. If $I$ is an answer set of $\PP$, then a weakly stable matching for $(S_M,S_W)$ is given by $\{(x,y) \,|\, accept(x,y)\in I\}$. Conversely, if $\{(x_{1},y_{1})$, $\hdots$, $(x_{k},y_{k})\}$ is a weakly stable matching for $(S_M,S_W)$ then~$\PP$ has the following answer set~$I$:
\allowdisplaybreaks
\begin{align*}
&\{manpropose(x_{i},y)\,|\, i \in \{1,\hdots,k\}, x_{i}\in M, y <_M^{x_i} y_{i} \vee y = y_i \neq x_i\} \\
\cup &\{womanpropose(x,y_{i})\,|\,i \in \{1,\hdots,k\}, y_{i}\in W, x <_W^{y_i} x_i \vee x = x_i \neq y_i\} \\
\cup &\{accept(x_{i},y_{i}) \,|\, i \in \{1,\hdots,k\}\} 
\end{align*} 
\end{proposition}
\begin{proof}
\nw{Let $(S_M,S_W)$ and $\PP$ be as described in the proposition. Because of the symmetry between the men and women we restrict ourselves to the male case when possible.}\\
\noindent \fbox{Answer set $\Ra$ weak stable set} 
We prove this in 4 steps.
\begin{enumerate}
\item \textit{For every $i\in \{1,\hdots,n\}$, every $j \in \{1,\hdots,p\}$ and for every answer set $I$ of $\PP$, it holds that $accept(m_i,w_j) \in I$ implies that $j \in acceptable_M^i$ and $i \in acceptable_W^j$.}\\
\nw{This can be proved by contradiction. We will prove that for every man $m_i$ and every $j \in unacceptable^i_M$, $accept(m_i,w_j)$ is in no answer set $I$ of the induced ASP program $\PP$. For $accept(m_i,w_j)$ to be in an answer set $I$, the reduct must contain some rule with this literal in the head and a body which is satisfied. The only rule for which this can be the case is the one of the form (1), implying that $manpropose(m_i,w_j)$ should be in $I$. But since $j$ is not in $acceptable^i_M$ there is no rule with $manpropose(m_i,w_j)$ in the head and so $manpropose(m_i,w_j)$ can never be in $I$.}
\item \textit{For every answer set $I$ of $\PP$ and every man $m_i$, there exists at most one woman $w_j$ such that $accept(m_i,w_j)\in I$. Similarly, for every woman $w_j$ there exists at most one man $m_i$ such that $accept(m_i,w_j)\in I$. Moreover, if $accept(m_i,m_i)\in I$ then $accept(m_i,w_j)\notin I$ for any $w_j$, and likewise when $accept(w_j,w_j)\in I$ then $accept(m_i,w_j)\notin I$ for any $m_i$.}\\
\nw{This can be proved by contradiction. Suppose first that there is an answer set $I$ of $\PP$ that contains $accept(m_i,w_j)$ and $accept(m_i,w_{j'})$ for some man $m_i$ and two different women $w_j$ and $w_{j'}$. The first step implies that $j$ and $j'$ are elements of $acceptable_M^i$. Either man $m_i$ prefers woman $w_j$ to woman $w_{j'}$ ($w_j \leq_M^{m_i} w_{j'}$), or the other way around ($w_{j'} \leq_M^{m_i} w_{j}$) or man $m_i$ has no preference among them ($w_j \leq_M^{m_i} w_{j'}$ and $w_{j'} \leq_M^{m_i} w_{j}$). The first two cases are symmetrical and can be handled analogously. The last case follows from the first case because it has stronger assumptions. We prove the first case and assume that man $m_i$ prefers woman $w_j$ to woman $w_{j'}$. The rules (4) imply the presence of a rule $manpropose(m_i,w_{j'})$ $\la \hdots$, $not \, accept(m_i,w_j), \hdots$ and this is the only rule which can make $manpropose(m_i,w_{j'})$ true (the only rule with this literal in the head). However, since $accept(m_i,w_j)$ is also in the answer set, the body of this rule is not satisfied so $manpropose(m_i,w_{j'})$ can never be in $I$. Consequently $accept(m_i,w_{j'})$ can never be in $I$ since the only rule with this literal in the head is of the form (1) and this body can never be satisfied, which leads to a contradiction.\\
Secondly assume that $accept(m_i,w_j)$ and $accept(m_i,m_i)$ are both in an answer set $I$ of $\PP$. Again step 1 implies that $j \in acceptable^i_M$. Because of the rules (2), $\PP$ will contain the rule $accept(m_i,m_i) \la$ $\hdots, not \, accept(m_i,w_j), \hdots$. An analogous reasoning as above implies that since $accept(m_i,w_j)$ is in the answer set $I$, $accept(m_i,m_i)$ can never be in $I$.}
\item \textit{For every man $m_i$, in every answer set $I$ of $\PP$ exactly one of the following conditions is satisfied}:
\begin{enumerate}
\item \textit{there exists a woman $w_j$ such that $accept(m_i,w_j) \in I$,}
\item $accept(m_i,m_i) \in I$,
\end{enumerate} 
\textit{and similarly for every woman $w_i$.} \\
\nw{Suppose $I$ is an arbitrary answer set of $\PP$ and $m_i$ is an arbitrary man. We already know from step 2 that a man cannot be paired to a woman while being single, so both possibilities are disjoint. Therefore, suppose there is no woman $w_j$ such that $accept(m_i,w_j)$ is in $I$. $\PP$ will contain the rule (2). Because of our assumptions and the definition of the reduct, this rule will be reduced to $accept(m_i,m_i) \la$, and so $accept(m_i,m_i)$ will be in $I$.}
\item \nw{For an arbitrary answer set $I$ of $\PP$ the previous steps imply that $I$ produces a set of marriages without blocking individuals. Weak stability also demands the absence of blocking pairs. Suppose by contradiction that there is a blocking pair $(m_i,w_j)$, implying that there exist $i\neq i'$ and $j \neq j'$ such that $accept(m_i,w_{j'}) \in I$ and $accept(m_{i'},w_j) \in I$ while $w_{j} <_M^{m_i} w_{j'}$ and $m_i <_W^{w_j} m_{i'}$. The rules of the form (1), which are the only ones with the literals $accept(m_i,w_{j'})$ and $accept(m_{i'},w_j)$ in the head, imply that literals $manpropose(m_i,w_{j'})$ and $womanpropose(m_{i'},w_j)$ should be in $I$. But since $w_{j} <_M^{m_i} w_{j'}$ and because of the form of the rules (4) there are fewer conditions to be fulfilled for $manpropose(m_i,w_j)$ to be in $I$ than for $manpropose(m_i,w_{j'})$ to be in $I$. Therefore, $manpropose(m_i,w_j)$ should be in $I$ as well. A similar reasoning implies that $womanpropose(m_i,w_j)$ should be in $I$. But now the rules of the form (1) imply that $accept(m_i,w_j)$ should be in $I$, contradicting step 2 since $accept(m_i,w_{j'})$ and $accept(m_{i'},w_j)$ are already in $I$.}
\end{enumerate}

\noindent \fbox{Weak stable set $\Ra$ answer set} 
\nw{Suppose we have a stable set of marriages $S=\{(x_1,y_1),\hdots,(x_k,y_k)\}$, implying that every $y_i$ is an acceptable partner of $x_i$ and the other way around. The rules of the form (1) do not alter when forming the reduct, but the other rules do as those contain naf-literals. Notice first that the stability of $S$ implies that there cannot be an unmarried couple $(m,w)$, with $m$ a man and $w$ a woman, such that $manpropose(m,w)$ is in $I$ and $womanpropose(m,w)$ is in $I$. By definition of $I$ this would mean that they both strictly prefer each other to their current partner in $S$. This means they would form a blocking pair, but since $S$ was stable, that is impossible. Therefore, the rules of the form (1) will be applied exactly for married couples $(m_i,w_j)$, since by definition of $I$ $manpropose(m_i,w_j)$ and $womanpropose(m_i,w_j)$ are both in $I$ under these conditions. For other cases the rule will also be fulfilled since the body will be false. This reasoning implies that the unique minimal model of the reduct w.r.t.\ $I$ should indeed contain $accept(m_i,w_j)$ for every married couple $(m_i,w_j)$ in $S$.
Since $S$ is a stable set of marriages, every person is either married or single. If a man $m_i$ is single, there will be no other literal of the form $accept(m_i,.)$ in $I$, so rule (2) will reduce to a fact $accept(m_i,m_i) \la$, which is obviously fulfilled by $I$. Similarly if a woman $w_j$ is single. Any other rule of the form (2) or (3) is deleted because $m_i$ or $w_j$ is not single in that case and thus there is some literal of the form $accept(m_i,w)$ for some woman $w$ and some literal of the form $accept(m,w_j)$ for some man $m$ in $I$, falsifying the body of the rules. If $m_i$ is single, then $accept(m_i,m_i)$ is in $I$ and this is the only literal of the form $accept(m_i,.)$ in $I$, so the rules of the form (4) will all be reduced to facts. The heads of these fact rules should be in the minimal model of the reduct and are indeed in $I$ since the women $w$ for which $manpropose(m_i,w)$ is in $I$ are exactly those who are strictly preferred to staying single. The rules of the form (4) for women $w_j$ in $neutral_M^i$ will all be deleted in this case, because $accept(m_i,m_i)$ is in $I$. If man $m_i$ is married to a certain woman $w_j$ in the stable set $S$ then the rules of the form (4) will reduce to facts of the form $manpropose(m_i,w) \la$ for every woman $w$ who is strictly preferred to $w_j$ and will be deleted for every other woman appearing in the head, because those rules will contain $not\, accept(m_i,w_j)$ in the body. Again $I$ contains these facts by definition, as the minimal model of the reduct should. We can use an analogous reasoning for the women.  The presence of the literals of the form $manpropose(.,.)$, $womanpropose(.,.)$ and $accept(.,.)$ in $I$ is thus required in the unique minimal model of the reduct w.r.t.\ $I$. We have proved that every literal in $I$ should be in the minimal model of the reduct and that every rule of the reduct is fulfilled by $I$, implying that $I$ is an answer set of $\PP$. }
\end{proof}

\section{Proof of Proposition 2} \label{sec:proof2}
\begin{proposition}
Let the criterion $crit$ be an element of $\{sexeq$, $weight$, $regret$, $singles\}$. For every answer set $I$ of the program $\PP_{crit}$ induced by an SMTI instance the set $S_I = \{(m,w) \,|\,$ $accept(m$, $w)$ $\in$ $I\}$ forms an optimal stable matching of marriages w.r.t.\ criterion $crit$ and the optimal criterion value is given by the unique value $v_I$ for which $crit(v_I) \in I$. Conversely for every optimal stable matching $S = \{(x_1,y_1), \hdots,$ $(x_k,y_k)\}$ with optimal criterion value $v$ there exists an answer set $I$ of $\PP_{crit}$ such that $\{(x,y) \,|\,$ $accept(x,y)$ $\in I\} = \{(x_i,y_i) \,|\, i \in \{1,\hdots,k\}\}$ and $v$ is the unique value for which $crit(v) \in I$.
\end{proposition}
\begin{proof}
Let $(S_M,S_W)$ be an SMTI instance. \\
\noindent \fbox{Answer set $\Ra$ optimal stable matching} Let $I$ be an arbitrary answer set of $\PP_{crit}$ and let $S_I$ be as formulated. It is clear that the only rules in $\PP_{crit}$ that influence the literals of the form $manpropose(.,.)$, $womanpropose(.,.)$ and $accept(.,.)$ are the rules in $\PP_{norm}$. Hence $I$ should contain an answer set $I_{norm}$ of $\PP_{norm}$ as a subset. Proposition 1 implies that $I_{norm}$ corresponds to a stable matching $S_I = \{(m,w) \,|\,$ $accept(m,w)$ $\in$ $I_{norm}\}$. Moreover, the only literals of the form $manpropose(.,.)$, $womanpropose(.,.)$ and $accept(.,.)$ in $I$ are those in $I_{norm}$, so $S_I = \{(m,w) \,|\,$ $accept(m,w)$ $\in$ $I\}$. If $crit=sexeq$, it is straightforward to see that the literals of the form $accept(.,.)$ in $I_{norm}$ uniquely determine which literals of the form $mancost(.,.)$, $womancost(.,.)$, $manweight(.)$, $womanweight(.)$ and $sexeq(.)$ should be in the answer set $I$. These literals do not occur in rules of $\PP_{crit}$ besides those in $\PP^{sexeq}_{ext}$. Note that the rules which do contain these literals imply that there will be just one literal of the form $sexeq(.)$ in $I$, namely $sexeq(v)$ with $v$ the sex-equality cost of $S_I$. Analogous results can be derived for $crit \in \{weight, regret, singles\}$. It remains to be shown that $S_I$ is an optimal stable matching. Suppose by contradiction that $S_I$ is not optimal, so there exists a stable matching $S^{\ast}$ such that $v_I > v^\ast$, with $v^\ast$ the criterion value of $S^\ast$ to be minimized. We prove that this implies that $I$ cannot be an answer set of $\PP_{crit}$, contradicting our initial assumption. \\
Proposition 1 and Lemma 1 imply that there exists an interpretation $I^\ast_{disj}$ of the ASP program $\PP_{disj}$ induced by $(S_M,S_W)$ that corresponds to the stable matching $S^\ast$. Moreover this interpretation is consistent, i.e.\ it will not contain $atom$ and $\neg atom$ for some atom $atom$. This implies that the interpretation $I'_{disj}$ defined as $I^\ast_{disj}$ in which $\neg atom$ is replaced by $natom$ for every atom $atom$ will falsify the body of the rules of the form (11) of $\PP'^{crit}_{ext}$. An analogous reasoning as above yields that the literals of the form $accept'(.,.)$ in $I'_{disj}$ uniquely determine which literals of the form $mancost'(.,.)$, $womancost'(.,.)$, $mansum'(.,.)$, $womansum'(.,.)$, $manweight'(.)$, $womanweight'(.)$ and $sexeq'(.)$ should be in $I'_{disj}$. With those extra literals added to $I'_{disj}$, we find that $I'_{disj}$ satisfies all the rules of $\PP'^{crit}_{ext}$. Moreover, $crit(v^\ast)$ is the unique literal of the form $crit(.)$ in $I'_{disj}$. Note that $I'_{disj}$ does not contain the atom $sat$.\\
Define the interpretation $J=I_{norm} \cup I'_{disj}$. From the previous argument it follows that $J$ will satisfy every rule of $\PP^{crit}_{ext} \cup \PP'^{crit}_{ext}$ since the predicates occurring in both programs do not overlap. Moreover $J$ contains $crit(v_I)$ and $crit'(v^\ast)$ and these are the only literals of the form $crit(.)$ or $crit'(.)$. Since $v_I > v^\ast$ the rules of the form (13) will be satisfied by $J$ since their body is always false. Call $J'$ the set $J \cup \{a \,|\, (a \la) \in \PP_{sat}\}$. Since $J'$ does not contain $sat$, the rules of $\PP_{sat}$ will all be satisfied by $J'$, with exception of the rule $\la not \, sat$. \\
The rule of the form (14) implies that $I$, as an answer set of $\PP_{crit}$, should contain $sat$. Now the set of rules (15) -- (16) imply that $I$ should also contain the literals $mancost'(.,.), womancost'(.,.)$ and $manpropose'(.,.)$, $womanpropose'(.,.)$, $accept'(.,.)$ with the corresponding literals prefixed by $n$ for every possible argument stated by the facts in $\PP_{sat}$. The rules (12), (21) and (23) in $\PP'^{crit}_{ext}$, by which we replaced rules (8) -- (9), (18) -- (19) and (22), guarantee that for every possible set of marriages and its corresponding criterion value $c$, $I$ will contain $crit(c)$ and all associated intermediate results. For example, for $crit=sexeq$, the rules will garantuee that $I$ also contains $mansum'(.,.)$, $manweight(.)$, $womansum(.,.)$ and $womanweight(.)$ for every argument that could occur in a model of $\PP'^{crit}_{ext}$. 
Note that this would not be the case if we used the original rules with $\#sum$, $\#max$ and $\#count$ in $\PP'^{crit}_{ext}$, since these rules would lead to only one value $c_M$ for which e.g.\ $manweight(c_M)$ should be in $I$, and similarly only one value $c_W$ for which $womanweight(c_W)$ should be in $I$. Consequently there would be only one value $c$ such that $crit(c)$ should be in $I$. This value would not necessarily correspond to $v^\ast$ and so we would not be able to conclude that $I'_{disj} \subseteq I$. 
However, with the current formulation of the rules we can conclude that $I'_{disj} \subseteq I$. We already reasoned in the beginning of the proof that $I_{norm} \subseteq I$ holds so it follows that $J \subseteq I$. Since the literals of $J' \setminus J$ are stated as facts of $\PP^{crit}_{ext}$, they should be in $I$, hence $J' \subseteq I$. Moreover $J' \subset I$ since $sat \in I \setminus {J'}$. \\
We use the notation $red(\PP,I)$ to denote the reduct of an ASP program $\PP$ w.r.t.\ an interpretation $I$. There is no rule in $\PP'^{crit}_{ext}$ with negation-as-failure in the body, hence $red(\PP'^{crit}_{ext},I) = red(\PP'^{crit}_{ext},J')$ = $\PP'^{crit}_{ext}$. We already reasoned that $J'$ satifies all the rules of the latter. We also reasoned that $I$ does not contain any other literals of the form $accept(.,.)$ than those which are also in $I_{norm}$, and by construction the same holds for $J'$. Hence $red(\PP^{crit}_{ext},I) = red(\PP^{crit}_{ext},J')$ and by construction $J'$ satisfies all the rules of this reduct. It is clear that $red(\PP_{sat},I)$ is $\PP_{sat}$ without the rule $\la not \, sat$, since $sat \in I$. Again we already argued that $J'$ satisfies $red(\PP_{sat},I)$. Hence $J'$ satisfies all the rules of $red(\PP_{crit},I)$, implying that $I$, which strictly contains $J'$, cannot be an answer set of $\PP_{crit}$ since it is not a minimal model of the negation-free ASP program $red(\PP_{crit},I)$~\cite{ASPGel88}. \\
\noindent \fbox{Optimal stable matching $\Ra$ answer set} 
\nw{Let $S = \{(x_1,y_1), \hdots, (x_k,y_k)\}$ be an optimal stable matching with optimal criterion value $v$. To see that the second part of the proposition holds it suffices to verify that the following interpretation $I$ is an answer set of $\PP_{crit}$, with the notation $P_{x_i}(y)$ as the index $a$ for which $y \in \sigma^l_M(a)$ if $x_i=m_l$ and symmetrically $P_{y_i}(x)$ as the index $a$ for which $x \in \sigma^{l'}_W(a)$ if $y_i=w_{l'}$. If $x_i=y_i$ we set $P_{x_i}(y_i)=P_{y_i}(x_i)=|\sigma^i_M|$ if $x_i$ is a man and $P_{x_i}(y_i)=P_{y_i}(x_i)=|\sigma^i_W|$ otherwise. So let $I$ be given by: $I = I_1 \cup I_2$ with }
\allowdisplaybreaks
\begin{align*}
I_1 = & \{accept (x_i,y_i) \,|\, i \in \{1,\hdots,k\}\} \cup \{crit(v)\} \cup \{sat\} \\
\cup& \{womanpropose(x_i,y_i) \,|\, x_i \neq y_i\} \{manpropose(x_i,y_i)|x_i \neq y_i\} \\
\cup& \{manpropose(x_i,y) \,|\, i \in \{1,\hdots,k\}, x_i=m_l, \exists a < P_{x_i}(y_i) \dpt y \in \sigma^l_M(a)\} \\
\cup& \{womanpropose(x,y_i)|i \in \{1,\hdots,k\}, y_i=w_{l'} ,\exists a < P_{y_i}(x_i) \dpt x \in \sigma^{l'}_W(a)\} \\
\cup& \{mancost(l,P_{x_i}(y_i)) \,|\, crit \neq singles, i \in \{1,\hdots,k\}, x_i = m_l\} \\ 
\cup& \{womancost(P_{y_i}(x_i),l') \,|\, crit \neq singles, i \in \{1,\hdots,k\}, y_i = w_{l'}\} \\
\cup& \{manweight(c_M(S)),womanweight(c_W(S)) \,|\, crit \in \{sexeq,weight\}\} \\
\cup& \{manregret(c_{regret,M}(S)),womanregret(c_{regret,W}(S))  \,|\, crit = regret\}
\end{align*}
and
\begin{align}
I_2 = &\{manargcost'_1(z) \,|\, 1 \leq z \leq n\} \cup  \{manargcost'_2(z) \,|\, 1 \leq z \leq p+1\} \nonumber\\
\cup & \{womanargcost'_1(z)  \,|\,1 \leq z \leq p\}\} \cup \{womanargcost'_2(z)  \,|\, 1 \leq z \leq n+1\} \nonumber\\
\cup & \{man(x) \,|\, x \in M\} \cup \{woman(x) \,|\, x \in W\} \label{eq:facts}\\
\cup &\{mancost'(i,j) \,|\, crit \neq singles, 1\leq i \leq n, 1\leq j \leq p+1\} \nonumber\\
\cup& \{womancost'(j,i) \,|\, crit \neq singles,1\leq i \leq n+1, 1\leq j \leq p\} \label{eq:mwcost}\\
\cup& \{manpropose'(x,y), womanpropose'(x,y) \,|\, x \in M, y \in W\} \nonumber\\
\cup &\{nmanpropose'(x,y), nwomanpropose'(x,y) \,|\, x \in M, y \in W\} \nonumber\\
\cup & \{accept'(x,y) \,|\, x \in M, y \in W\} \cup \{accept'(x,x) \,|\,x \in M \cup W\}  \nonumber\\
\cup &\{naccept'(x,y) \,|\, x \in M, y \in W\} \cup \{naccept'(x,x) \,|\,x \in M \cup W\} \label{eq:mpwpacc}\\
\cup& \{crit'(val) \,|\, val \in \arg(crit)\} \nonumber\\
\cup & \{single'(i,j)\,|\, crit=singles, 1 \leq i \leq n+p, j \in \{0,1\}\}\nonumber\\
\cup & \{singlesum'(i,j)\,|\, crit=singles, 1 \leq i \leq n+p, 1 \leq j \leq n+p-i+1\}\}\nonumber\\
\cup& \{mansum'(i,j) \,|\, crit \in \{sexeq,weight\}, 1 \leq i \leq n,\nonumber \\
&\hspace{160pt} n-i+1 \leq j \leq (n-i+1)(p+1)\} \nonumber\\
\cup& \{womansum'(j,i) \,|\, crit \in \{sexeq,weight\}, 1 \leq j \leq p,\nonumber\\
&\hspace{160pt} p-j+1 \leq i \leq (p-i+1)(n+1)\} \nonumber\\
\cup& \{manweight'(z) \,|\, crit \in \{sexeq,weight\}, n \leq z \leq n(p+1)\} \nonumber\\
\cup& \{womanweight'(z) \,|\, crit \in \{sexeq,weight\}, p \leq z \leq p(n+1)\} \nonumber\\
\cup& \{manmax'(i,j) \,|\, crit = regret, 1 \leq i \leq n,1 \leq j \leq p+1\} \nonumber\\
\cup& \{womanmax'(j,i) \,|\, crit = regret, 1 \leq j \leq p, 1\leq i \leq n+1\} \nonumber\\
\cup& \{manregret'(z) \,|\, crit=regret, 1 \leq z \leq p+1\} \nonumber\\
\cup& \{womanregret'(z) \,|\, crit=regret,1\leq z \leq n+1\} \label{eq:rest}
\end{align}  
The notation $\arg(c)$ stands for the possible values the criterion can take within this problem instance:
\begin{itemize}
\item if $crit=sexeq$ then $\arg(crit)=\{0,\hdots, \max(np+n-p,np+p-n)\}$,
\item if $crit=weight$ then $arg(crit)=\{n+p,\hdots,2np+p+n\}$,
\item if $crit=regret$ then $\arg(crit)=\{1,\hdots,\max(p,n)+1\}$,
\item if $crit=singles$ then $\arg(crit)=\{0,\hdots,n+p\}$.
\end{itemize}
\nw{To verify whether this interpretation is an answer set of $\PP_{crit}$, we should compute the reduct w.r.t.\ $I$ and check whether $I$ is a minimal model of the reduct~\cite{ASPGel88}. It can readily be checked that $I$ satisfies all the rules of $red(\PP_{crit},I)$. It remains to be shown that there is no strict subset of $I$ which satisfies all the rules. First of all, all the facts of $\PP_{crit}$ must be in the minimal model of the reduct, explaining why the sets of literals (\ref{eq:facts}) should be in $I$. The only rules with negation-as-failure are part of $\PP^{crit}_{ext}$.\\
As in the previous part of the proof, it is straightforward to see that $I_1$ is the unique minimal model of the reduct of $\PP^{crit}_{ext}$ w.r.t.\ $I$, considering that the literals in $I_2$ do not occur in $\PP^{crit}_{ext}$. So any minimal model of $red(\PP_{crit},I)$ must contain $I_1$.\\
The key rule which makes sure that $I$ is a minimal model of the reduct is (13). The rules (11) imply that for each model of $red(\PP_{crit},I)$ that does not contain $sat$, the literals of $\PP'^{crit}_{ext}$ in that model will correspond to a stable matching of the SMTI instance. In that case rule (13) will have a true body, since $S$ is optimal, implying that $sat$ should have been in the model. And the presence of $sat$ in any minimal model implies the presence of the set of literals (\ref{eq:rest}) in any minimal model of the reduct. This can be seen with the following reasoning. Due to the presence of the facts (\ref{eq:facts}) and $sat$ in any minimal model of the reduct, rules (15) imply the presence of the literals (\ref{eq:mwcost}) in any minimal model. For the same reason rules (16) imply that the literals (\ref{eq:mpwpacc}) should be in any minimal model of $red(\PP_{crit},I)$. For $crit=sexeq$ the presence of the literals of the form (\ref{eq:mwcost}) in any minimal model of the reduct together with rules (12) imply that $mansum'(i,j)$ should be in any minimal model for every $i \in \{1,\hdots,n\}$ and $j \in \{n-i+1,\hdots,(n-i+1)(p+1)\}$: for $i=n$, the first rule of (12) implies that $mansum'(n,x)$ is in any minimal model for every $x$ such that $manargcost_2'(x)$ is in it, i.e.\ any $x \in \{1,\hdots,p+1\}$. Now the second rule of (12) implies that $mansum'(n-1,x)$ is in any minimal model for every $x+y$ such that $manargcost_2'(x)$ and $mansum'(n,y)$ are in it, i.e.\ any $x+y \in \{2,\hdots,2(p+1)\}$. If we continue like this, it is straightforward to see that every literal of the form $mansum'(.,.)$ of $I_2$ should be in any minimal model. The third rule of (12) now implies that $manweight'(x)$ should be in any minimal model for every $x$ such that $mansum'(1,x)$ is in it, i.e.\ $x \in \{n,\hdots,n(p+1)\}$. The same reasoning can be repeated for the literals $womansum'$ and $womanweight'$. At this point rules (10) imply that $sexeq'(|x-y|)$ should be in any minimal model which contains $manweight'(x)$ and $womanweight'(y)$. Note that only one of the two rules in (10) will apply for every $x$ and $y$ since the numerical variables in DLV are positive. Considering the arguments for which $manweight'$ and $womanweight'$ should be in any minimal model, it follows that $sexeq'(x)$ should be in any minimal model for every $x \in \{0,\hdots,\max(p(n+1)-n,n(p+1)-p)\}$, which is exactly $\arg(crit)$. For the other criteria, an analogous reasoning shows that the presence of all literals of $I_2$ is required in any minimal model of the reduct.\\
Considering the fact that we have proved that all literals of $I$ should be in any minimal model of the reduct and $I$ fulfils all the rules of the reduct, we know that $I$ is a minimal model of the reduct and thus an answer set of $\PP_{crit}$.}
\end{proof}
\end{appendix}

\bibliographystyle{acmtrans}
\bibliography{biblio}